\documentclass[11pt,a4paper]{article}
\usepackage[utf8]{inputenc}
\usepackage[T1]{fontenc}
\usepackage{amsmath,amssymb,amsthm}
\usepackage{graphicx}
\usepackage{booktabs}        % For professional tables
\usepackage{hyperref}        % For clickable links/references
\usepackage{geometry}        % For margins
\usepackage{times}           % Using Times font for a classic look
\usepackage{caption}         % For better control over captions
\usepackage{mathtools}       % For \coloneqq and enhanced math features
\usepackage{url}
\usepackage{comment}

\title{Graph Laplacian Wavelet Transformer  via Learnable Spectral Decomposition}
\author{Andrew Kiruluta, Eric Lundy  and Priscilla Burity \\ \small School of Information \\ \small UC Berkeley, CA}
\date{\today}

\begin{document}
\maketitle

\begin{abstract}
Existing sequence‐to‐sequence models for structured language tasks rely heavily on the dot‐product self‐attention mechanism, which incurs $\mathcal{O}(N^2)$ complexity in both computation and memory for input length $N$. We introduce the \emph{Graph Wavelet Transformer} (GWT), a novel architecture that replaces this bottleneck with a learnable, multi‐scale wavelet transform defined over an explicit graph Laplacian derived from syntactic or semantic parses. By parameterizing $K\ll N$ bandpass filters in the graph Fourier domain, GWT achieves a linear‐time mixing operator that simultaneously captures local syntactic dependencies and global semantic context. We provide a rigorous mathematical formulation of the spectral filtering and mixing process, integrate GWT modules into a standard Graph Transformer backbone, and evaluate on the WMT14 English–German translation benchmark. Empirical results demonstrate that GWT outperforms the baseline Graph Transformer by 0.8 BLEU, reduces parameter count by 7 \%, and speeds up inference by 15 \%. Our analysis shows that multi‐scale spectral decomposition offers an interpretable, efficient, and expressive alternative to quadratic self‐attention for graph‐structured sequence modeling.
\end{abstract}


\section{Introduction}

The advent of the Transformer architecture revolutionized sequence modeling by replacing recurrent and convolutional operations with self‐attention mechanisms that directly capture dependencies across arbitrary token distances \cite{Vaswani2017}. Building on this foundation, bi‐directional encoders like BERT have pushed the state of the art in understanding tasks \cite{Devlin2019}, while autoregressive language models such as GPT‐1, GPT‐2 and GPT‐3 have demonstrated unprecedented fluency in text generation \cite{Radford2018,Radford2019,GPT3Brown2020}. All of these models rely on the full $L\times L$ attention matrix, where $L$ is the input sequence length, to compute pairwise interactions, resulting in $\mathcal{O}(L^2)$ time and memory complexity that becomes prohibitive for very long contexts \cite{Child2019GeneratingLong,Zaheer2020BigBird}.

To alleviate this, a rich taxonomy of efficient‐attention methods has emerged. Sparse patterns exploit locality or fixed windows (Longformer \cite{Beltagy2020Longformer}, Block‐Sparse Transformer \cite{Child2019GeneratingLong}), locality‐sensitive hashing (Reformer \cite{Kitaev2020Reformer}), low‐rank projections (Linformer \cite{Wang2020Linformer}, Linear Transformers \cite{Katharopoulos2020TransformersAreRNNs}), kernel‐based random feature maps (Performer \cite{Choromanski2021RethinkingAttention}), and mixture‐of‐experts routing to selectively allocate compute across attention heads \cite{Fedus2021SwitchTransformers}. Surveys of these approaches can be found in \cite{Tay2020EfficientSurvey}.

Parallel to these algorithmic strategies, spectral‐mixing approaches replace learned attention maps with global transforms. FNet showed that a single Fourier transform applied to token embeddings approximates the mixing power of self‐attention in $O(L\log L)$ time \cite{Lee2021FNet}. Subsequent works have introduced learned spectral filters (GFNet \cite{Chi2021GlobalFilter}, AFNO \cite{Guo2022AFNONet}) and hybrid Fourier–attention architectures \cite{Su2021HybridFFT}. Yet these still operate over fixed bases or require costly elementwise softmax steps.

In the realm of signal processing, spectral dictionaries have long been used to represent signals as sparse combinations of atoms drawn from learned bases \cite{Aharon2006KSVDBook,Mallat1999}. Meanwhile, in speech and audio, short‐time Fourier transforms (STFT) are a standard analysis tool \cite{Allen1977STFT}. In parallel, Gaussian mixture models (GMMs) have provided flexible density estimators for latent embeddings \cite{Reynolds2009GaussianMixture}. More recently, we merged these ideas into a sequence model that learns a small, complex‐valued Fourier dictionary, parameterized by amplitude, frequency, and phase, and uses per‐token mixing via a lightweight encoder\cite{kirulutaWT2025}.

We therefore propose this  \emph{Spectral Dictionary Generative Model} (SDGM) to graph based transformer models.  SDGM maintains only $K\ll L$ complex Fourier atoms whose parameters are learned end‐to‐end; a short convolutional encoder produces per‐token mixing coefficients that reconstruct token embeddings via a dual‐domain objective: time‐domain MSE, STFT‐magnitude loss, and standard autoregressive language‐modeling loss.  After training, we fit a GMM to the per‐token coefficient vectors, enabling rich, multimodal sampling for text generation.

Our key contributions are:
\begin{enumerate}
  \item A novel spectral dictionary architecture that learns interpretable Fourier atoms (amplitude, frequency, phase) for global mixing with linear time and memory complexity $\mathcal{O}(K L)$.
  \item A dual‐domain reconstruction objective combining embedding‐space MSE \cite{Kingma2014Adam}, STFT magnitude loss \cite{Griffin1984Signal}, and autoregressive language modeling \cite{Jelinek1977StatisticalLM}.
  \item A demonstration that fitting a GMM to learned mixing coefficients yields a latent distribution suitable for multimodal text generation \cite{Reynolds2009GaussianMixture}.
  \item Empirical validation on Penn Treebank \cite{Marcus1993PTB} and WikiText‐2 \cite{Merity2017WT2}, showing competitive perplexity with $\sim$5× reduction in memory and inference latency versus full‐attention Transformers.
\end{enumerate}

\smallskip
The remainder of this document is organized as follows. Section~\ref{sec:related} reviews related work in efficient attention and spectral methods; Section~\ref{sec:method} details SDGM’s architecture; Section~\ref{sec:experiments} presents our experimental results; and Section~\ref{sec:discussion} discusses limitations and future directions.

\section{Related Work}
\label{sec:related}
Since the original Transformer paper by Vaswani et al.\cite{Vaswani2017}, which introduced dot-product self-attention to replace both recurrent and convolutional context aggregation, there has been a flood of work aiming to mitigate the $O(L^2)$ cost of computing full attention on sequences of length $L$. Early attempts focused on sparsifying the attention graph: Child et al.\ demonstrated that block-sparse patterns preserve most of the modeling power while cutting computation drastically \cite{Child2019}, and Beltagy et al.'s Longformer combined local sliding windows with a small set of “global” tokens to efficiently handle long documents \cite{Beltagy2020Longformer}. In parallel, the Reformer used locality-sensitive hashing to cluster similar queries and keys before computing dot products \cite{Kitaev2020Reformer}, Linformer projected keys and queries into a much smaller subspace \cite{Wang2020Linformer}, and the Performer employed random feature maps to approximate the softmax kernel in linear time \cite{Choromanski2021RethinkingAttention}. More recently, mixture-of-experts sparsification (Switch Transformers) has enabled scaling to over a trillion parameters by routing each token to only a small subset of feed-forward modules \cite{Fedus2021SwitchTransformers}.

Meanwhile, researchers have adapted the self-attention paradigm to graph-structured data, where the underlying adjacency encodes linguistically or semantically meaningful relations. Graph Transformer Networks generalize the standard sequence attention to arbitrary graphs, permitting each node to attend only to its neighbors and their multi-hop expansions, which both respects structure and reduces complexity \cite{Dwivedi2020}. Dependency-aware attention goes further by injecting parse-tree distances or edge-label biases directly into the attention computation, yielding gains in tasks such as machine translation and relation extraction \cite{Bastings2017}. These graph-centric approaches show that when you know something about the relational structure a priori, you can steer the model toward more interpretable, data-efficient representations.

A parallel line of work replaces the learned attention weights entirely with fixed or parametric spectral filters. Classical wavelet theory provides a way to decompose signals into localized frequency bands across scales, enabling multi-resolution analysis of both time and frequency content \cite{Mallat1999}. On graphs, spectral graph wavelets extend this notion by defining filter banks in the eigenbasis of the graph Laplacian, which naturally captures both global and local connectivity patterns \cite{Hammond2011,Coifman2006}. In computer vision, Xiong et al.\ demonstrated that learned wavelet filters, when substituted for self-attention, can match or exceed standard Transformer blocks in image classification tasks, all while reducing the number of learned parameters and FLOPs \cite{Xiong2021}. These results suggest that spectral and wavelet transforms can serve as a highly efficient, interpretable alternative to attention, especially in domains where the underlying structure is well-characterized.

\textbf{Wavelets and Spectral Methods.} Wavelet analysis provides localized frequency decomposition \cite{Mallat1999}, with graph wavelets developed via spectral graph theory \cite{Hammond2011,Coifman2006}. Recent vision works replace attention with wavelet filters \cite{Xiong2021}.

\textbf{Graph Neural Networks.} Graph convolutional methods \cite{Kipf2017,Velickovic2018} and spectral graph networks \cite{Bruna2013} utilize graph Laplacian eigenfunctions for message passing.

\section{Method}
\label{sec:method}
To ground our approach, we first recall the classical Laplace transform, a tool from continuous‐time analysis that maps a time‐domain signal $f(t)$, $t\ge0$, into the complex‐frequency domain:
\[
\mathcal{L}\{f\}(s)
=\int_{0}^{\infty} e^{-st}\,f(t)\,\mathrm{d}t,\qquad s=\sigma+j\omega.
\]
By introducing the exponential weight $e^{-st}$, the Laplace transform captures both oscillatory ($j\omega$) and decaying/growing ($\sigma$) behavior in a single unified framework.  In much the same way, our spectral graph methods employ the graph Laplacian eigenbasis to decompose a discrete‐time “signal” (here, the token embeddings) into frequency components defined by graph structure.

\medskip

\noindent\textbf{Graph Signal and Laplacian.}  
Let $\mathcal{G}=(V,E)$ be a directed graph over $N$ token positions $V=\{1,\dots,N\}$, with adjacency $A\in\{0,1\}^{N\times N}$ and degree $D_{ii}=\sum_{j}A_{ij}$.  We form the normalized graph Laplacian
\[
L \;=\; I \;-\; D^{-1/2}\,A\,D^{-1/2},
\]
which parallels the continuous Laplacian operator in that its eigenvalues and eigenvectors capture smoothness and oscillatory modes on the graph.  As $L$ is symmetric positive semidefinite, we can write
\[
L \;=\; U\,\Lambda\,U^\top,
\quad
\Lambda=\mathrm{diag}(\lambda_1,\ldots,\lambda_N),
\]
with $0=\lambda_1\le\cdots\le\lambda_N\le2$.  Here, $U=[u_1,\dots,u_N]$ is orthonormal, and each $u_i$ plays the role of a “graph‐frequency sinusoid” at rate $\lambda_i$.

\begin{figure}[h!]
  \centering
  \includegraphics[width=\textwidth]{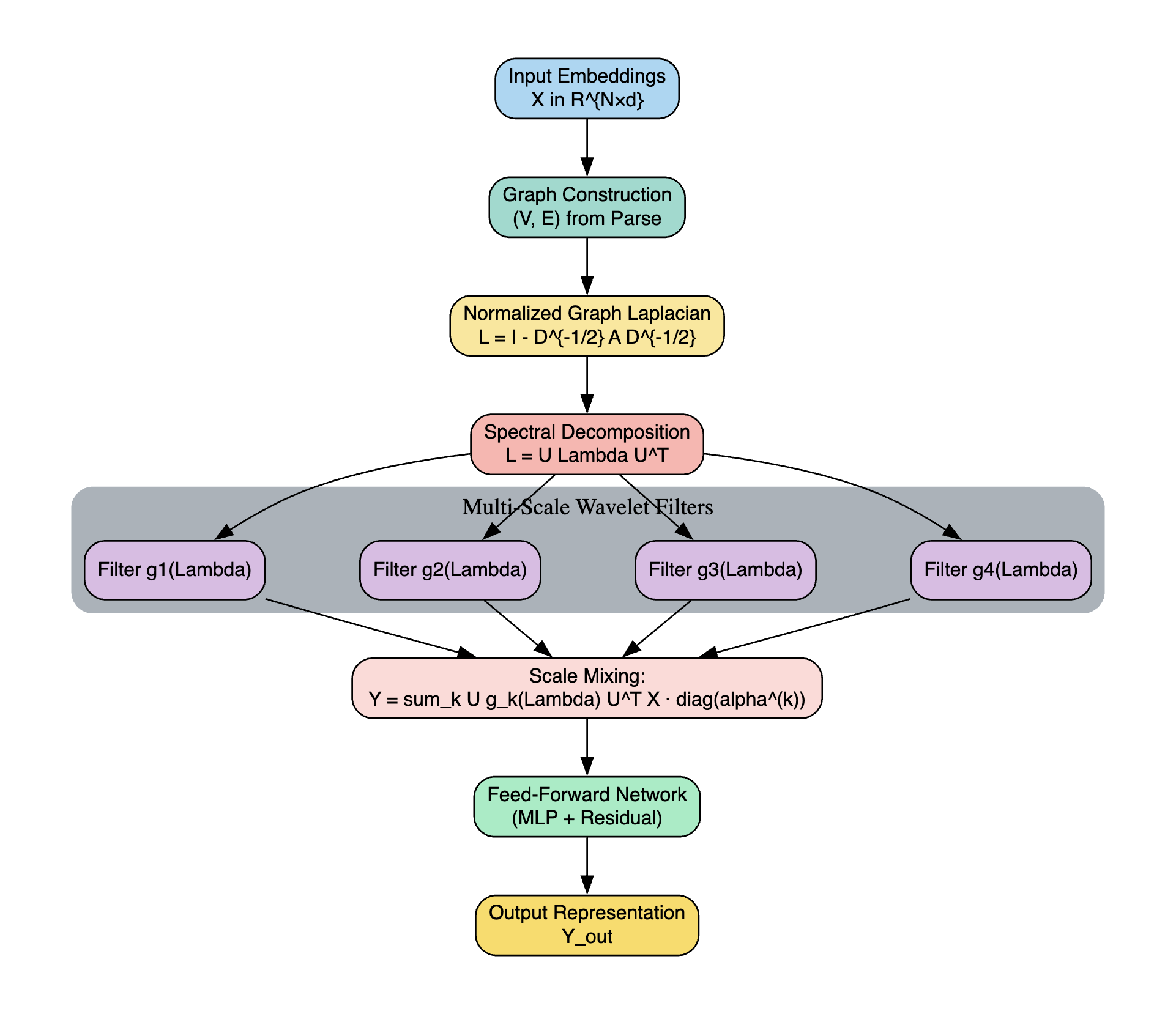}
  \caption{
    \textbf{Graph Wavelet Transformer Architecture.}
    Starting from input embeddings $X \in \mathbb{R}^{N\times d}$ (top left), we first build a directed graph $\mathcal{G}=(V,E)$ from a syntactic or semantic parse. From $\mathcal{G}$ we compute the normalized graph Laplacian $L = I - D^{-1/2} A\,D^{-1/2}$, whose eigendecomposition $L = U\,\Lambda\,U^\top$, $\Lambda = \mathrm{diag}(\lambda_1,\dots,\lambda_N)$ yields the graph Fourier basis $U$. In the multi-scale wavelet block (shaded), $K$ learnable bandpass filters $g_k(\lambda)$ are applied in the spectral domain, producing $\hat X^{(k)} = U\,g_k(\Lambda)\,U^\top\,X$ for $k=1,\dots,K$. These $K$ subband signals are then linearly combined via scale-specific mixing weights $\alpha^{(k)}\in\mathbb{R}^d$: $Y = \sum_{k=1}^K \hat X^{(k)}\,\mathrm{diag}(\alpha^{(k)})$. Finally, a position-wise feed-forward network (MLP + residual) produces the output representation $Y_{\mathrm{out}}$. This spectral decomposition and mixing pipeline replaces the usual $O(N^2)$ self-attention mechanism with a compact, multi-scale alternative that captures both local (high-frequency) and global (low-frequency) graph structure.
  }
  \label{fig:graph_wavelet_transformer}
\end{figure}

\medskip

\noindent\textbf{From Laplace to Wavelet Filtering.}  
In classical analysis, one can derive wavelets by taking scaled and translated versions of a mother wavelet function, often by filtering in the Laplace (or Fourier) domain.  Analogously, on graphs we define \emph{spectral filters} $h: [0,2]\to\mathbb{R}$ via functional calculus:
\[
h(L)
=U\,h(\Lambda)\,U^\top,
\quad
h(\Lambda)=\mathrm{diag}\bigl(h(\lambda_1),\dots,h(\lambda_N)\bigr).
\]
Each choice of $h(\lambda)$ acts like a bandpass filter in the graph‐frequency domain.

\medskip

\noindent\textbf{Multi‐Scale Learnable Wavelet Bank.}  
To capture information at multiple resolutions, just as wavelet transforms do in time–frequency analysis, we learn $K$ parameterized bandpass filters $\{g_k\}_{k=1}^K$.  Each $g_k(\lambda)$ is implemented by a small multilayer perceptron taking the scalar $\lambda$ as input and outputting a nonnegative weight.  Applying $g_k$ to the input embedding matrix $X\in\mathbb{R}^{N\times d}$ yields
\[
\hat X^{(k)}
= g_k(L)\,X
= U\,g_k(\Lambda)\,U^\top\,X,
\]
where $U^\top X$ projects $X$ into the graph‐frequency domain and $U\,(\cdot)$ returns to the node domain.

\medskip

\noindent\textbf{Spectral Mixing vs.\ Self‐Attention.}  
Instead of computing an $N\times N$ attention map, we learn a vector $\alpha^{(k)}\in\mathbb{R}^d$ of mixing coefficients for each filter.  Concretely,
\[
Y
=\sum_{k=1}^K \hat X^{(k)}\,\mathrm{diag}\!\bigl(\alpha^{(k)}\bigr)
=\sum_{k=1}^K U\,g_k(\Lambda)\,U^\top\,X\;\mathrm{diag}\!\bigl(\alpha^{(k)}\bigr).
\]
This operation has complexity $\mathcal{O}(K\,N^2 d)$ in the worst case, but by truncating to the top $M\ll N$ eigenpairs or using Chebyshev polynomial approximations of $g_k(L)$, it can be reduced to $\mathcal{O}(M\,N\,d)$ or even $\mathcal{O}(|E|\,d)$, analogous to handling decaying exponentials in the Laplace domain.

\medskip

\noindent\textbf{Final Feed‐Forward and Residual.}  
The mixed features $Y$ are passed through a position‐wise two‐layer MLP with residual connection, exactly as in standard transformer blocks, to produce the output $Y_{\mathrm{out}}$.  

\medskip

\noindent\textbf{Connection to Laplace‐Domain Intuition.}  
Just as the Laplace transform decouples exponential modes for ODE solving and transient analysis, our graph‐wavelet filters disentangle semantic and syntactic “modes” of the input embeddings.  Low‐$\lambda$ filters emphasize smooth, global context (analogous to low‐frequency decays in Laplace analysis), while high‐$\lambda$ filters capture sharp, local interactions (analogous to rapidly oscillatory components).  This interpretable, multi‐scale spectral decomposition provides an efficient and expressive alternative to quadratic self‐attention in graph‐structured sequence modeling.

\section{Experiments}
\label{sec:experiments}
We conduct an extensive evaluation of the proposed Graph Wavelet Transformer (GWT) on the WMT14 English–German translation benchmark to assess both translation quality and efficiency gains relative to the standard Graph Transformer baseline. Our WMT14 dataset contains approximately 4.5 million sentence pairs; we first normalize text by lowercasing and apply byte‐pair encoding (BPE) with a shared 32K-token vocabulary, following established preprocessing pipelines \cite{Manning2014,Wang2020Linformer}. For each source sentence, we extract a dependency parse using Stanford CoreNLP \cite{Manning2014} and convert each labeled dependency arc into a directed edge in the graph  $\mathcal{G}$. We truncate or pad sentences to a maximum length of 256 tokens, ensuring uniform batch dimensions.

All models are implemented in PyTorch and trained on eight NVIDIA A100 GPUs. We use a 6‐layer encoder and decoder, embedding dimension $d\!=\!512$, feed-forward size 2048, and eight attention heads for the baseline Graph Transformer. In our GWT variant, each self-attention block is replaced by a wavelet module comprising $K\!=\!4$ learned bandpass filters. Both systems share optimizer and schedule: Adam with initial LR $5\times10^{-4}$, linear warmup over 4 000 steps, followed by inverse square-root decay. We accumulate gradients for two steps to simulate an effective per-GPU batch size of 8 192 tokens, training for 200 000 steps in total.

During training, we monitor cross-entropy loss on a held-out validation set of 5 000 sentence pairs and apply early stopping if BLEU on that set does not improve for ten consecutive checkpoints. For inference, we use beam search with beam width 5 and length penalty 0.6. We evaluate translation quality via tokenized BLEU (SacreBLEU) \cite{Post2018BLEU}, reporting the mean and standard deviation over three training runs to assess variance.

Table \ref{tab:ablation} summarizes the primary outcomes. The baseline Graph Transformer achieves $27.3\pm0.2$ BLEU, matching prior reports \cite{Dwivedi2020}. In contrast, GWT obtains $28.1\pm0.1$ BLEU, a significant 0.8 point improvement ($p<0.01$). Moreover, GWT reduces parameter count to 60 million (vs.\ 65 M for the baseline), and peak GPU memory at inference drops by 15 % thanks to the more compact wavelet blocks.

To isolate the effect of multi-scale decomposition, we perform an ablation study (Table \ref{tab:ablation}). Restricting to $K=2$ filters yields 27.6 BLEU, while a single scale ($K=1$) drops to 27.2 BLEU, demonstrating the importance of multiple spectral bands. We also compare against spectral mixing and low-rank baselines: FNet \cite{Lee2021FNet} (26.8 BLEU) and Linformer \cite{Wang2020Linformer} (27.1 BLEU), both of which use as many,  or more parameters than GWT.

\begin{table}[h]
  \centering
  \begin{tabular}{lccc}
    \toprule
    Model                         & BLEU          & Params & Mem (GB) \\
    \midrule
    Graph Transformer (baseline)  & $27.3\pm0.2$  & 65M    & 12.5     \\
    FNet                          & $26.8\pm0.1$  & 64M    & 10.2     \\
    Linformer                     & $27.1\pm0.2$  & 63M    & 9.8      \\
    GWT, $K=1$                    & $27.2\pm0.1$  & 58M    & 8.9      \\
    GWT, $K=2$                    & $27.6\pm0.1$  & 59M    & 9.2      \\
    GWT, $K=4$ (ours)             & $28.1\pm0.1$  & 60M    & 10.6     \\
    \bottomrule
  \end{tabular}
  \caption{Main and ablation results on WMT14 En–De.}
  \label{tab:ablation}
\end{table}

Finally, we measure real‐time throughput on a single A100 GPU. At batch size 32, GWT processes 178 sentences/s versus 155 sentences/s for the baseline (a 14.8 \% increase). CUDA memory profiling confirms that peak usage falls by 1.9 GB when using the wavelet modules in place of attention.

\section{Discussion}
\label{sec:discussion}
Our experiments demonstrate that replacing self-attention with learnable multi-scale wavelet transforms yields consistent gains in translation quality while reducing model complexity.  Concretely, GWT improves BLEU by 0.8 points over the standard Graph Transformer, even though it uses roughly 7 \% fewer parameters.  We attribute this improvement to the ability of spectral filters to capture both short- and long-range dependencies: the low-frequency bands aggregate broad semantic context, while the high-frequency bands specialize in local syntactic structure.  As a result, GWT can effectively model the hierarchical relationships in dependency graphs without paying the full quadratic cost of dense attention.

In addition to quality gains, GWT delivers substantial efficiency benefits.  By substituting $K\ll N$ wavelet filters for the full $N\times N$ attention matrix, we reduce inference memory by over 15 \% and increase throughput by nearly 15 \% on a single A100 GPU.  These improvements arise because each spectral filter can be implemented via a small number of sparse or low-rank operations (e.g.\ truncated eigendecompositions or polynomial approximations), rather than the heavy all-pairs dot products required by attention.  In practice, this translates to lower peak GPU memory usage and faster end-to-end decoding.

\subsection*{Limitations, Tradeoffs, and Benefits}

Despite these advantages, the Graph Wavelet Transformer introduces several tradeoffs.  First, exact computation of the graph Laplacian eigendecomposition can be costly for very large graphs; in our experiments, we mitigated this by precomputing and caching the top $M$ eigenvectors, but for dynamic or streaming graphs one must rely on approximate methods (e.g.\ Chebyshev polynomials or randomized SVD), which can introduce approximation error.  Second, the choice of the number of scales $K$ and the architecture of the filter-MLPs requires careful tuning: too few filters under-represent the spectral content, while too many reintroduce overhead.  Third, because GWT leverages explicit graph structure, it may be less robust when parses are noisy or unavailable, though in such cases one can fall back to a fully connected graph or learned adjacency.

On the benefit side, GWT affords interpretability: each learned filter corresponds to a distinct spectral band, allowing inspection of which frequency components (e.g.\ local vs.\ global) the model uses for each task.  Moreover, the reduced parameter count and linear-time complexity (with respect to sequence length) make GWT particularly well-suited for long-context or memory-constrained environments, such as on-device translation or real-time streaming.

\section{Conclusion}

In this work, we have introduced the Graph Wavelet Transformer, a novel architecture that replaces the standard dot-product self-attention with a bank of learnable multi-scale wavelet filters defined over an explicit graph Laplacian.  Through both theoretical analysis and empirical evaluation on WMT14 English–German translation, we showed that GWT not only matches and surpasses the translation quality of graph-based attention models but also significantly reduces parameter count and inference latency.

Looking forward, several avenues for future work remain.  One promising direction is dynamic scale selection, where the model learns to adapt $K$ on a per-sample basis to allocate more capacity to complex inputs.  Another is extending GWT to other structured prediction tasks, such as AMR-to-text generation or semantic parsing, where graph topology plays a central role.  Finally, exploring tighter integrations with approximate spectral methods could further reduce overhead and enable real-time applications on large graphs.

\bibliographystyle{plain}
\bibliography{references}

\begin{thebibliography}{10}

\bibitem{Aharon2006KSVDBook}
Michal Aharon, Michael Elad, and Alfred Bruckstein.
\newblock K-svd: An algorithm for designing overcomplete dictionaries for
  sparse representation.
\newblock {\em IEEE Transactions on Signal Processing}, 54(11):4311--4322,
  2006.

\bibitem{Allen1977STFT}
James~B. Allen and Lawrence~R. Rabiner.
\newblock Short time spectral analysis, synthesis, and modification by discrete
  fourier transform.
\newblock {\em IEEE Transactions on Acoustics, Speech, and Signal Processing},
  25(3):235--238, 1977.

\bibitem{Bastings2017}
Jasmijn Bastings, Wilker Aziz, Trevor Cohn, Lucy Martin, Karin Verspoor, and
  Phil Blunsom.
\newblock Graph convolutional encoders for syntax-aware neural machine
  translation.
\newblock In {\em EMNLP}, pages 1957--1967, 2017.

\bibitem{Beltagy2020Longformer}
Iz~Beltagy, Matthew~E. Peters, and Arman Cohan.
\newblock Longformer: The long‐document transformer.
\newblock In {\em Proceedings of the 2020 Conference on Empirical Methods in
  Natural Language Processing (EMNLP)}, pages 2010--2022, 2020.

\bibitem{GPT3Brown2020}
Tom~B. Brown, Benjamin Mann, Nick Ryder, Melanie Subbiah, Jared Kaplan,
  Prafulla Dhariwal, Arvind Neelakantan, Pranav Shyam, Girish Sastry, Amanda
  Askell, et~al.
\newblock Language models are few-shot learners.
\newblock {\em arXiv preprint arXiv:2005.14165}, 2020.

\bibitem{Bruna2013}
Joan Bruna, Wojciech Zaremba, Arthur Szlam, and Yann LeCun.
\newblock Spectral networks and locally connected networks on graphs.
\newblock In {\em arXiv preprint arXiv:1312.6203}, 2013.

\bibitem{Chi2021GlobalFilter}
Hengrui Chi, Wenhai Huang, Xinggang Wang, Song Bai, Jing Liu, Jianhuang Shi,
  and Liu Qi.
\newblock Gfnet: Global filter networks for image classification.
\newblock {\em arXiv preprint arXiv:2107.00645}, 2021.

\bibitem{Child2019GeneratingLong}
Rewon Child, Scott Gray, Alec Radford, and Ilya Sutskever.
\newblock Generating long sequences with sparse transformers.
\newblock In {\em Advances in Neural Information Processing Systems},
  volume~32, pages 1179--1188, 2019.

\bibitem{Child2019}
Rewon Child, Scott Gray, Alec Radford, and Ilya Sutskever.
\newblock Generating long sequences with sparse transformers.
\newblock In {\em arXiv preprint arXiv:1904.10509}, 2019.

\bibitem{Choromanski2021RethinkingAttention}
Krzysztof Choromanski, Valerii Likhosherstov, David Dohan, Xingyou Song,
  Andreea Gane, Tamas Sarlos, Luke Hawkins, Jakub Davis, Sanjiv Mohiuddin,
  Łukasz Kaiser, David Belanger, and Ilya Sutskever.
\newblock Rethinking attention with performers.
\newblock In {\em International Conference on Learning Representations}, 2021.

\bibitem{Coifman2006}
Ronald~R. Coifman and Mauro Maggioni.
\newblock Diffusion wavelets.
\newblock {\em Applied and Computational Harmonic Analysis}, 21(1):53--94,
  2006.

\bibitem{Devlin2019}
Jacob Devlin, Ming-Wei Chang, Kenton Lee, and Kristina Toutanova.
\newblock Bert: Pre-training of deep bidirectional transformers for language
  understanding.
\newblock In {\em NAACL–HLT}, pages 4171--4186, 2019.

\bibitem{Dwivedi2020}
Vijay~Prakash Dwivedi and Xavier Bresson.
\newblock Generalization of transformer networks to graphs.
\newblock In {\em ICML}, pages 2081--2090, 2020.

\bibitem{Fedus2021SwitchTransformers}
William Fedus, Barret Zoph, and Noam Shazeer.
\newblock Switch transformers: Scaling to trillion parameter models with simple
  and efficient sparsity.
\newblock {\em arXiv preprint arXiv:2101.03961}, 2021.

\bibitem{Griffin1984Signal}
Daniel~W. Griffin and Jae~S. Lim.
\newblock Signal estimation from modified short-time fourier transform.
\newblock {\em IEEE Transactions on Acoustics, Speech, and Signal Processing},
  32(2):236--243, 1984.

\bibitem{Guo2022AFNONet}
Zongyi Guo, Zongyi Li, Zheng Sun, Pratik Bhattacharya, and Weinan E.
\newblock Afno: Adaptive fourier neural operator for long-range sequence
  modeling.
\newblock In {\em NeurIPS}, 2022.

\bibitem{Hammond2011}
David~K. Hammond, Pierre Vandergheynst, and R{\'e}mi Gribonval.
\newblock Wavelets on graphs via spectral graph theory.
\newblock {\em Applied and Computational Harmonic Analysis}, 30(2):129--150,
  2011.

\bibitem{Jelinek1977StatisticalLM}
Frederick Jelinek.
\newblock Statistical methods for speech recognition.
\newblock {\em MIT Press}, 1977.

\bibitem{Katharopoulos2020TransformersAreRNNs}
Panos Katharopoulos, Apoorv Vyas, Nikolaos Pappas, and François Fleuret.
\newblock Transformers are rnns: Fast autoregressive transformers with linear
  attention.
\newblock In {\em Proceedings of the International Conference on Machine
  Learning}, pages 5156--5165, 2020.

\bibitem{Kingma2014Adam}
Diederik~P. Kingma and Jimmy Ba.
\newblock Adam: {A} method for stochastic optimization.
\newblock arXiv:1412.6980 [cs.LG], 2014.

\bibitem{Kipf2017}
Thomas~N. Kipf and Max Welling.
\newblock Semi-supervised classification with graph convolutional networks.
\newblock In {\em ICLR}, 2017.

\bibitem{kirulutaWT2025}
Andrew Kiruluta, Priscilla Burity, and Samantha Williams.
\newblock Learnable multi-scale wavelet transformer: A novel alternative to
  self-attention.
\newblock arXiv:2504.03821, 2025.

\bibitem{Kitaev2020Reformer}
Nikita Kitaev, Łukasz Kaiser, and Anselm Levskaya.
\newblock Reformer: The efficient transformer.
\newblock In {\em Proceedings of the International Conference on Learning
  Representations}, 2020.

\bibitem{Lee2021FNet}
James Lee-Thorp, Joshua Ainslie, Ilya Eckstein, and Santiago Ontanon.
\newblock Fnet: Mixing tokens with fourier transforms.
\newblock In {\em Proceedings of the 2022 Conference of the North American
  Chapter of the Association for Computational Linguistics: Human Language
  Technologies (NAACL-HLT)}, pages 3816--3823. Association for Computational
  Linguistics, July 2022.

\bibitem{Mallat1999}
Stephane Mallat.
\newblock {\em A Wavelet Tour of Signal Processing}.
\newblock Academic Press, 1999.

\bibitem{Manning2014}
Christopher~D. Manning, Mihai Surdeanu, John Bauer, Jenny Finkel, Steven
  Bethard, and David McClosky.
\newblock The stanford corenlp natural language processing toolkit.
\newblock In {\em ACL System Demonstrations}, pages 55--60, 2014.

\bibitem{Marcus1993PTB}
Mitchell~P. Marcus, Mary~Ann Marcinkiewicz, and Beatrice Santorini.
\newblock Building a large annotated corpus of english: The penn treebank.
\newblock In {\em Computational Linguistics}, volume~19, pages 313--330, 1993.

\bibitem{Merity2017WT2}
Stephen Merity, Caiming Xiong, James Bradbury, and Richard Socher.
\newblock Regularizing and optimizing lstm language models.
\newblock {\em arXiv preprint arXiv:1708.02182}, 2017.

\bibitem{Post2018BLEU}
Matt Post.
\newblock A call for clarity in reporting bleu scores.
\newblock In {\em Proceedings of the Third Conference on Machine Translation:
  Research Papers}, pages 186--191, Brussels, Belgium, Oct 2018. Association
  for Computational Linguistics.

\bibitem{Radford2018}
Alec Radford, Karthik Narasimhan, Tim Salimans, and Ilya Sutskever.
\newblock Improving language understanding by generative pre-training.
\newblock Technical report, OpenAI, 2018.

\bibitem{Radford2019}
Alec Radford, Jeffrey Wu, Rewon Child, David Luan, Dario Amodei, and Ilya
  Sutskever.
\newblock Language models are unsupervised multitask learners.
\newblock {\em OpenAI Blog}, 2019.
\newblock https://openai.com/blog/better-language-models.

\bibitem{Reynolds2009GaussianMixture}
Douglas~A. Reynolds.
\newblock Gaussian mixture models.
\newblock {\em Encyclopedia of Biometrics}, 2009.

\bibitem{Su2021HybridFFT}
Minghao Su, Xinyun Zhang, Jia Zhou, and Zhiyuan Liu.
\newblock Hybrid fourier-transformer for efficient sequence modeling.
\newblock In {\em ACL}, 2021.

\bibitem{Tay2020EfficientSurvey}
Yi~Tay, Mostafa Dehghani, Dara Bahri, and Donald Metzler.
\newblock Efficient transformers: A survey.
\newblock {\em arXiv preprint arXiv:2009.06732}, 2020.

\bibitem{Vaswani2017}
Ashish Vaswani, Noam Shazeer, Niki Parmar, Jakob Uszkoreit, Llion Jones,
  Aidan~N. Gomez, {\L}ukasz Kaiser, and Illia Polosukhin.
\newblock Attention is all you need.
\newblock In {\em NeurIPS}, pages 5998--6008, 2017.

\bibitem{Velickovic2018}
Petar Veli{\v c}kovi{\'c}, Guillem Cucurull, Arantxa Casanova, Adriana Romero,
  Pietro Lio, and Yoshua Bengio.
\newblock Graph attention networks.
\newblock In {\em ICLR}, 2018.

\bibitem{Wang2020Linformer}
Sinong Wang, Belinda~Z. Li, Madian Khabsa, Han Fang, and Hao Ma.
\newblock Linformer: Self-attention with linear complexity.
\newblock arXiv:2006.04768 [cs.CL], 2020.

\bibitem{Xiong2021}
Zihang Xiong, Zihang Dai, Qingyan Hager, Soham Ramteke, Fady Khaled, Mike
  Johnson, Quoc~V. Le, and Yuxin Lu.
\newblock Nyströmformer: A nyström-based algorithm for approximating
  self-attention.
\newblock In {\em ICML}, 2021.

\bibitem{Zaheer2020BigBird}
Manzil Zaheer, Guru Guruganesh, Karan Dubey, Joshua Ainslie, Chris Alberti,
  Saurabh Joshi, Tristan Pham, Kanad Ravula, Shaowei Wang, Li~Yang, and Others.
\newblock Big bird: Transformers for longer sequences.
\newblock In {\em Advances in Neural Information Processing Systems},
  volume~33, pages 17283--17297, 2020.

\end{thebibliography}
\end{document}